\documentclass[conference]{IEEEtran}
\IEEEoverridecommandlockouts
\usepackage{cite}
\usepackage{amsmath,amssymb,amsfonts}
\usepackage{algorithmic}
\usepackage{graphicx}
\usepackage{textcomp}
\usepackage{xcolor}
\usepackage{booktabs}
\usepackage[normalem]{ulem}
\usepackage{subcaption}
\useunder{\uline}{\ul}{}

\def\BibTeX{{\rm B\kern-.05em{\sc i\kern-.025em b}\kern-.08em
    T\kern-.1667em\lower.7ex\hbox{E}\kern-.125emX}}
\begin{document}

\title{Leveraging AV1 motion vectors for Fast and Dense Feature Matching  \\
\thanks{This work was funded by the Horizon CL4 2022, EU Project Emerald, 101119800.}
}

\author{\IEEEauthorblockN{Julien Zouein, Hossein Javidnia, Fran\c{c}ois Piti\'e, Anil Kokaram}
\IEEEauthorblockA{
Sigmedia Group, \\
Department of Electronic and Electrical Engineering, \textit{Trinity College Dublin}, Dublin, Ireland \\
\{zoueinj, hossein.javidnia, pitief, anil.kokaram\}@tcd.ie
}
}
\maketitle

\begin{abstract}

We repurpose AV1 motion vectors to produce dense sub-pixel correspondences and short tracks filtered by cosine consistency. On short videos, this compressed-domain front end runs comparably to sequential SIFT while using far less CPU, and yields denser matches with competitive pairwise geometry. As a small SfM demo on a 117-frame clip, MV matches register all images and reconstruct 0.46–0.62M points at 0.51–0.53\,px reprojection error; BA time grows with match density. These results show compressed-domain correspondences are a practical, resource-efficient front end with clear paths to scaling in full pipelines.

\end{abstract}


\section{Introduction}
Classical vision pipelines expend substantial computation on feature extraction and exhaustive matching, often consuming the majority of wall-clock time and CPU resources. Meanwhile, modern video codecs (e.g., AV1) embed motion information and block structure that can be repurposed for correspondence discovery at a fraction of the cost. Given that most digital video is stored and transmitted in a compressed format, with motion vectors required for decoding and readily available in the bitstream, re-using these pre-computed motion vectors is an inexpensive proxy for correspondence discovery. Prior work has explored this direction: in 2014, Kantorov et al.~\cite{kantorov} showed that motion vectors from H.264 could be used to speed up action recognition. More recently, Richard N.~C. Turner ~\cite{mov-slam} explored a pipeline for Simultaneous Localization and Mapping (SLAM) from motion vectors extracted from the H.265 bitstream.

We present an efficient approach that converts AV1 motion vectors (MVs) into sub-pixel point-to-point correspondences and builds \textit{multi-frame tracks}, unlocking long-baseline coverage. We evaluate the resulting correspondences with pairwise geometric tests and efficiency metrics, and include a small SfM demonstration; comprehensive 3D evaluation and BA scaling are left to a forthcoming longer paper.

The main contributions of this paper are:

i) Sub-pixel correspondence extraction from AV1 MVs via precise target placement.

ii) Track propagation and cosine filtering to extend coverage to non-adjacent pairs and prune outliers, producing a triangular image adjacency pattern.

iii) Pairwise geometry and efficiency protocol reporting inlier ratio, median Sampson error, and pre-stage runtime / CPU load; BA only in a small demo.

\section{Method}

In this section, we provide an overview of Motion Vectors in AV1 and how to extract them. Next, the structure of our proposed framework is described in detail.

\subsection{Converting Motion Vectors into Sub-pixel Point-to-Point Correspondences}\label{mv_av!}

The motion estimation process in modern hybrid video codecs like AV1 and H.265 relies on a shared block-based prediction architecture. In this paradigm, a frame is partitioned into blocks of various dimensions; AV1 supports a notably wide range from (4$\times$4) to (128$\times$128). The prediction for each block is generated by referencing a motion vector (MV) which points to a source location in a designated Reference Frame. The AV1 specification accommodates up to seven such reference frames~\cite{av1:adaptivepred, zhao2021tool} which can be configured for forward (predicting from future frames) or backward prediction. Our implementation focuses on a backward-predictive streaming configuration, where all motion vectors refer to locations in previously decoded frames.

To improve prediction accuracy, AV1 employs a high-precision motion model. The motion vectors are defined with a sub-pixel precision up to 1/8th of a pixel; in our experiments, we use 1/4th pixel precision. Pixel precision is achieved by generating an interpolated sub-pixel grid using separable filters~\cite{technical_overview_av1}. This allows for a more finely-grained and accurate motion compensation. Within the bitstream itself, these motion vectors are encoded as integer values for compactness. To convert these raw integer values back to their true physical scale upon extraction, a division by a factor of 8 is required.

Some blocks are encoded using INTRA prediction or are designated as SKIPPED blocks. These blocks do not have any associated motion vectors, resulting in a (0,0) motion vector during extraction. We do not consider blocks with a (0,0) motion vector, as this value is ambiguous and can represent either a truly static block or a block with no motion information.

For each block (p,q) in a frame $n$, we emit a source keypoint at the center of the block and a \textit{target point} displaced by the motion vector $v_{n,m}(p,q)$ in the reference frame $m$. The generated target point is added to the source keypoints of frame $m$.

\subsection{Track Propagation and Cosine Filtering}

We build tracks by linking consistent correspondences across consecutive frames and discard short tracks (length $<3$). Denote by $\mathbf{v}_{i,j}(k)\!\in\!\mathbb{R}^2$ the motion vector for keypoint $k$ from frame $i$ to frame $j$, and by $t_k=\{(n,\mathbf{x}_n),(m,\mathbf{x}_m),(\ell,\mathbf{x}_\ell),\dots\}$ the ordered set of detections (with $n<m<\ell<\dots$) that forms the track for $k$.

Codec motion vectors are block-prediction signals rather than true object motion, so some vectors are unreliable. We therefore enforce directional consistency within each track using a cosine test between adjacent MV segments. For any consecutive triple $(n,m,\ell)\in t_k$ we require
\begin{equation}
\label{cosine_filtering}
\frac{\langle \mathbf{v}_{n,m}(k),\,\mathbf{v}_{m,\ell}(k)\rangle}
     {\|\mathbf{v}_{n,m}(k)\|\,\|\mathbf{v}_{m,\ell}(k)\|} \;\ge\; 1-\epsilon,
\end{equation}
with a tolerance $\epsilon\in(0,1)$. By default we use $\epsilon{=}0.1$ (i.e., $\cos\!\ge\!0.9$); for sequences with large temporal gaps we disable the filter by setting $\epsilon{=}1$ (threshold $0$). If either vector has magnitude below a small $\tau$, we skip the test to avoid numerical instability.

\begin{figure}
    \centering
    \begin{tabular}{cc}
      \includegraphics[width=0.4\linewidth]{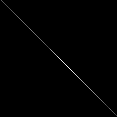}   & 
      \includegraphics[width=0.4\linewidth]{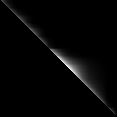}
      \\
      (a) & (b)
    \end{tabular}
    \caption{Visualisation of the impact of using tracks on the adjacency matrix. (a): adjacency before using tracks, (b): adjacency after using tracks and cosine filtering. For a given row $i$ and column $j$, the whiter a pixel is, the more matches there are.}
    \label{adjacency_evolution}
\end{figure}

Using cosine similarity in tracks allows us to filter out motion vectors with a bad direction. Motion vectors with bad cosine similarity are deleted and not considered for matches.

This processing step allows us to propagate good matches to non-adjacent frames, augmenting the number of matches. This transforms the diagonal adjacency (adjacent pair only for video and sequential matching) into a \textit{triangular} pattern with broad-baseline coverage as shown on Figure~\ref{adjacency_evolution}.

\begin{figure}[t]
    \centering
    \includegraphics[width=\columnwidth]{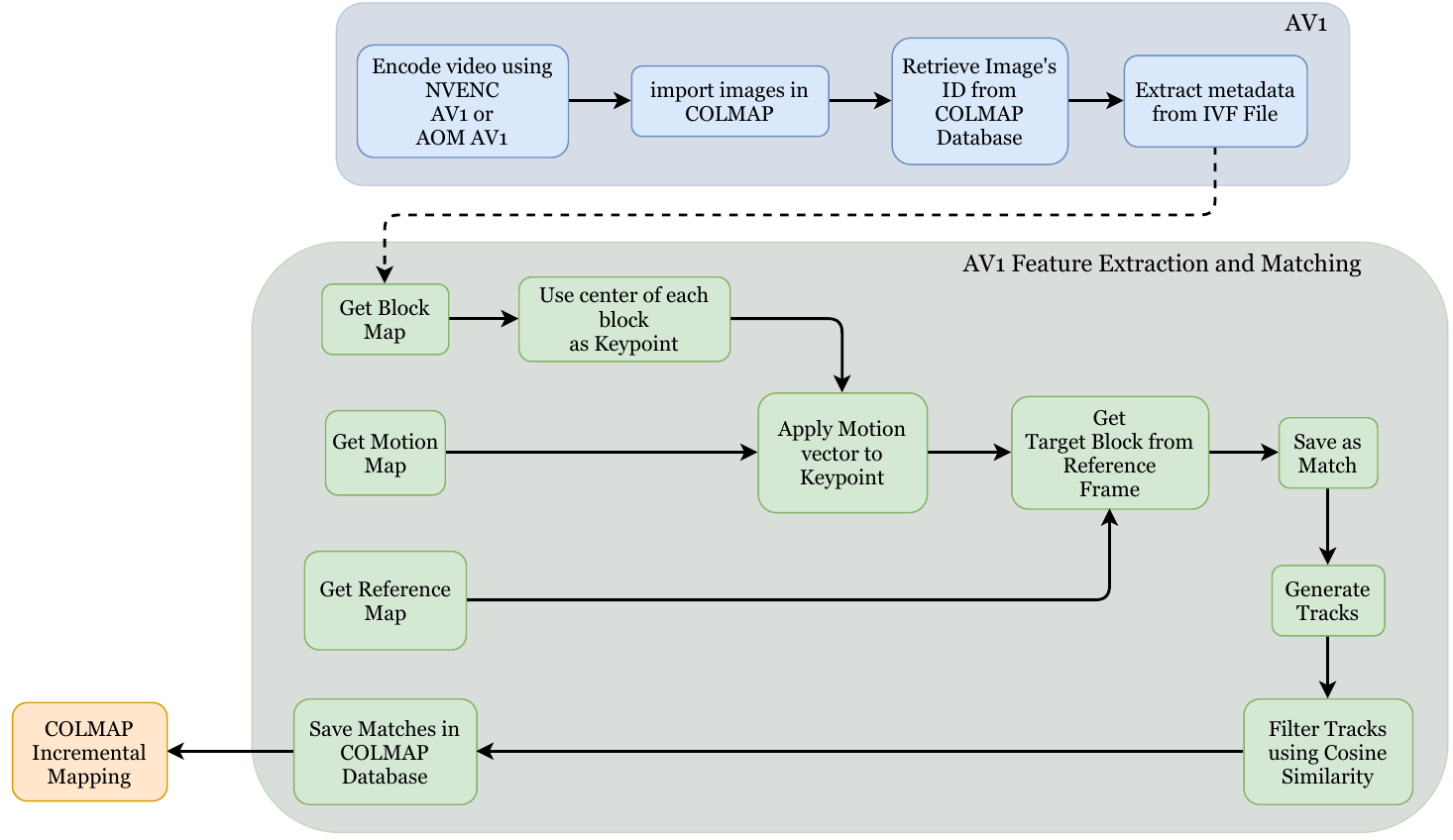}
    \caption{Pipeline from AV1 bitstream to correspondences and tracks. We parse block/motion/reference maps, generate sub-pixel point-to-point matches, build and filter tracks and export correspondences. Downstream SfM/BA is not the focus; we include a small demo in §III-C.}
    \label{fig:AV1_sfm_pipeline}
\end{figure}

Figure~\ref{fig:AV1_sfm_pipeline} is a representation of our current pipeline.

\subsection{Pairwise Geometry Estimation and Scoring}\label{metrics_description}
For each image pair $(i,j)$ we estimate a calibrated essential matrix $\mathbf{E}_{ij}$ or, for near-planar cases, a homography $\mathbf{H}_{ij}$ using (LO-)RANSAC with the fixed hyperparameters from §\emph{Baseline and settings}. Points are expressed in normalized camera coordinates $\tilde{\mathbf{x}}=K^{-1}\mathbf{x}$. For $\mathbf{E}$, the residual is the Sampson error~\cite{Sampson}:
\[
SE(x,x')=\frac{(x'^\top \mathbf{E}x)^2}{(\mathbf{E}x)_1^2+(\mathbf{E}x)_2^2+(\mathbf{E}^\top x')_1^2+(\mathbf{E}^\top x')_2^2},
\]
which approximates the epipolar distance. We fit both $\mathbf{E}$ (five-point) and $\mathbf{H}$ (DLT) and select the model with the larger inlier set, breaking ties by lower median residual (favoring $\mathbf{E}$ unless $\mathbf{H}$ explains substantially more matches).

\textbf{Scoring.} We report (i) the RANSAC inlier ratio and (ii) the median Sampson error over inliers. All metrics use the same correspondences and identical RANSAC settings across methods.

\section{Experiments}

We evaluate our method against traditional baselines using the datasets and settings described below. Performance is measured along three dimensions: pairwise geometry quality as defined in Section~\ref{metrics_description}, computational efficiency (runtime, CPU usage), and correspondence coverage (match count). We do not include the time for the final SfM reconstruction, as our goal is to evaluate the front-end pipeline.

\subsection{Experimental Setup}

We run experiments on a 12th Gen Intel(R) Core(TM) i7-12700K with 64GB RAM, running Ubuntu 22.04. The hardware encoding was performed using an NVIDIA RTX 6000 Ada.

\subsection{Dataset}

We evaluate our pipeline using outdoor sequences. Our test data includes \textit{Gerrard Hall}, and \textit{Person Hall} from COLMAP~\cite{colmap} collection of datasets. It is important to note that we use a subset of each dataset. Our technique requires images to have the same dimensions and to be temporally adjacent. These two image sets were converted into videos to meet our method's requirement for sequential inputs. Due to the large temporal gaps, we disable the cosine filter for these two sequences by setting $\epsilon=1$ (threshold 0) to preserve track continuity.

We also recorded three custom sequences: \begin{itemize}
    \item Dublin Seq. 1: having 329 images and a resolution of 1080$\times$1920 at 24 FPS, recorded at night.
    \item Paris Seq. 1: having 117 frames and a resolution of 1080$\times$1920 at 10 FPS, recorded in daylight.
    \item Paris Seq. 2: having 122 frames and a resolution of 1920$\times$1080 at 10 FPS, recorded in daylight.
\end{itemize}
All three sequences were recorded using an iPhone 15 Pro Max.
We added the first 230 frames of the Sequence 0 from KITTI Odometry dataset\cite{kitti_odometry}.

\subsection{SfM}
\begin{figure}
    \centering
    \begin{tabular}{c}
      \includegraphics[width=0.7\linewidth]{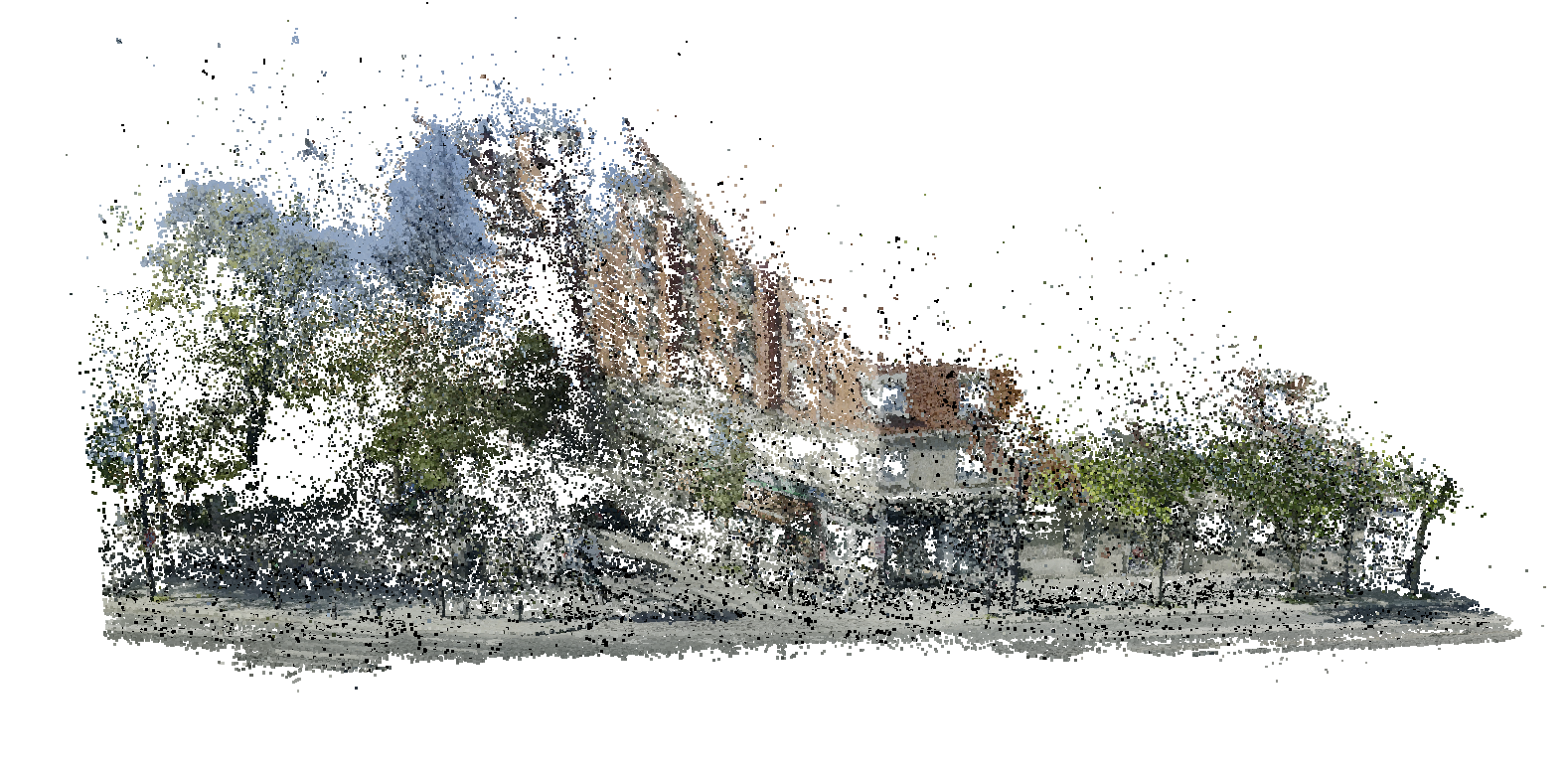} \\
    \end{tabular}
    \caption{3D reconstruction of Sequence Paris Seq 1. using MV (NVENC).
    }
    \label{SfM-exemple}
\end{figure}

Quantitative results. Table \ref{tab:sfm} reports SfM statistics on a 117-frame clip using sequential exhaustive pairing and identical SIMPLE\_RADIAL intrinsics and mapper settings across all methods. Our motion-vector pipeline (MV) registers all 117 images and reconstructs substantially more 3D points (460k–616k) than feature-matching baselines, with competitive reprojection error (0.51–0.53 px). The front-end time of MV is moderate (122–257 s); the larger BA time is expected because BA scales with the number of tracks/points. SIFT-Seq attains the lowest reprojection error (0.30 px) but yields an order-of-magnitude fewer points. GPU deep matchers (DISK~\cite{disk}+LightGlue~\cite{lightglue} and SP~\cite{superpoint}+SuperGlue~\cite{superglue}) run on GPU (marked *), but are either much slower or reconstruct very few points under identical mapping settings. Figure~\ref{SfM-exemple} shows the result of 3D reconstruction using our MV (NVENC).

\begin{table}[t]
\centering
\caption{SfM on a short video ($n{=}117$ frames). Times are seconds; \emph{front-end} = features+matching+geometric verification, \emph{BA} = mapping/BA. GPU methods marked $^{*}$ was performed using a NVIDIA T4.}
\resizebox{\columnwidth}{!}{%
\begin{tabular}{lcccc}
\toprule
Method & Reg. imgs & \#3D pts & Reproj. err (px) & Time (front-end / BA) \\
\midrule
MV (ours AOM)       & 117 & 460{,}152  & 0.51    & 122 / 1035 \\
MV (ours, NVENC)    & 117 & 615{,}945     & 0.53    & 257 / 1262 \\
SIFT-Seq            & 117 & 54{,}755     & 0.30   & 114 / 346 \\
DISK$^{*}$+LightGlue$^{*}$ & 117 & 85{,}351 & 1.07 & 1067 / 1495 \\
SP$^{*}$+SuperGlue$^{*}$  & 117 & 28{,}589 & 1.34 & 1320 / 412  \\
\bottomrule
\end{tabular}
}
\label{tab:sfm}
\end{table}

\subsection{Encoder Parameters}

For AV1 encoding, we employed libaom-av1~\cite{libaom} (v3.12.1) as the software encoder, and FFmpeg~\cite{ffmpeg} (n.6.0-22) to access the NVIDIA AV1 hardware implementation (nvenc). Motion vectors from AV1-encoded IVF files were extracted using the inspect tool from AOM library.

Encoders are set up in a Streaming Configuration (S3-SCC-03~\cite{av1_stream_conf}) where all the reference frames are \textit{in the past}, with only one intra-frame (first frame of the sequence) and all motion vectors pointing to the previous frame. We use preset 1 for NVENC-AV1 and cpu-used=6 for libaom.

\subsection{Baseline and Settings}

We use COLMAP SIFT~\cite{SIFT} with exhaustive matching (all pairs) and also report sequential matching as an alternative.

For geometric verification and model estimation, we apply RANSAC with fixed hyperparameters across all evaluated methods to maintain consistency.
We perform multiple independent runs with RANSAC and report median performance metrics over these repeated runs to reduce the impact of outliers. We estimate $\mathbf{E}$ with the five-point algorithm and RANSAC with default parameters (max\_error = 4.0, min\_inlier\_ratio=0.25, max\_num\_trials=10000). Sampson error is computed on normalized coordinates.

Regarding cosine similarity filtering, we set $\epsilon=0.1$ in equation~\ref{cosine_filtering}, to ensure a cosine similarity above 0.9 except for \textit{Person Hall} and \textit{Gerrard Hall} sequences, where the high temporal difference between each frame is not compatible with this filtering.

\subsection{Main Qualitative Results}

\begin{figure*}[t]
\centering
\begin{subfigure}[t]{0.22\textwidth}
    \centering
    \includegraphics[width=\linewidth]{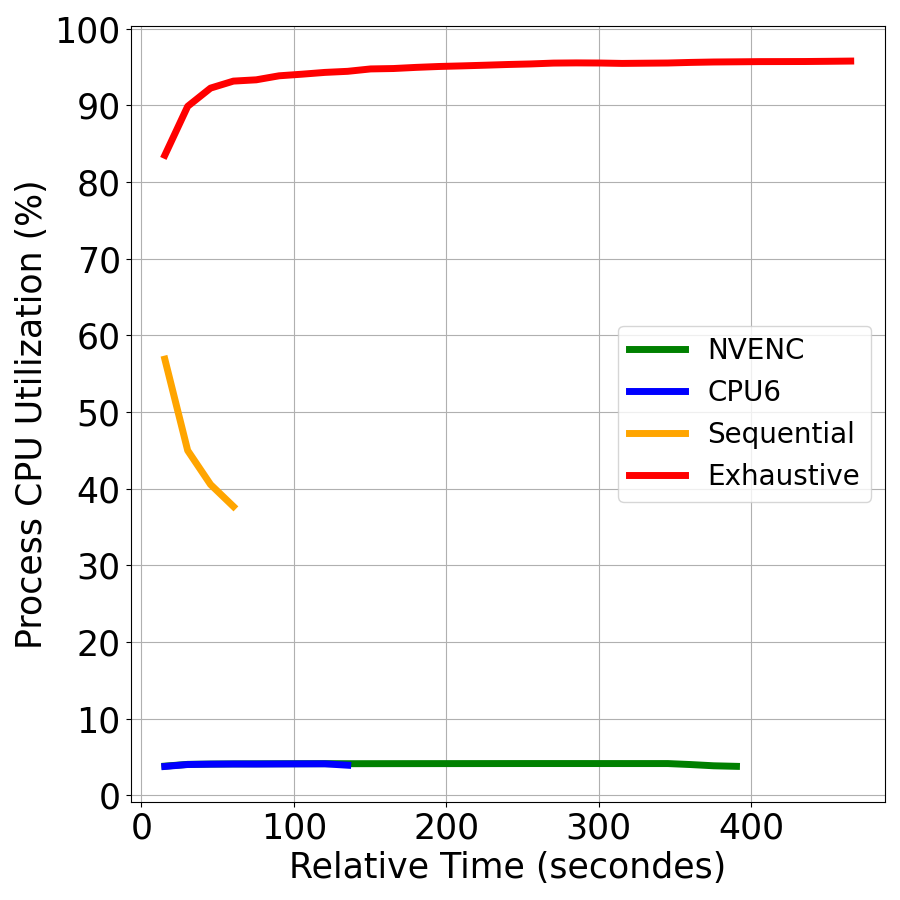}
    \caption{KITTI-00: CPU}
    \label{fig:cpu-kitti}
\end{subfigure}\hfill
\begin{subfigure}[t]{0.22\textwidth}
    \centering
    \includegraphics[width=\linewidth]{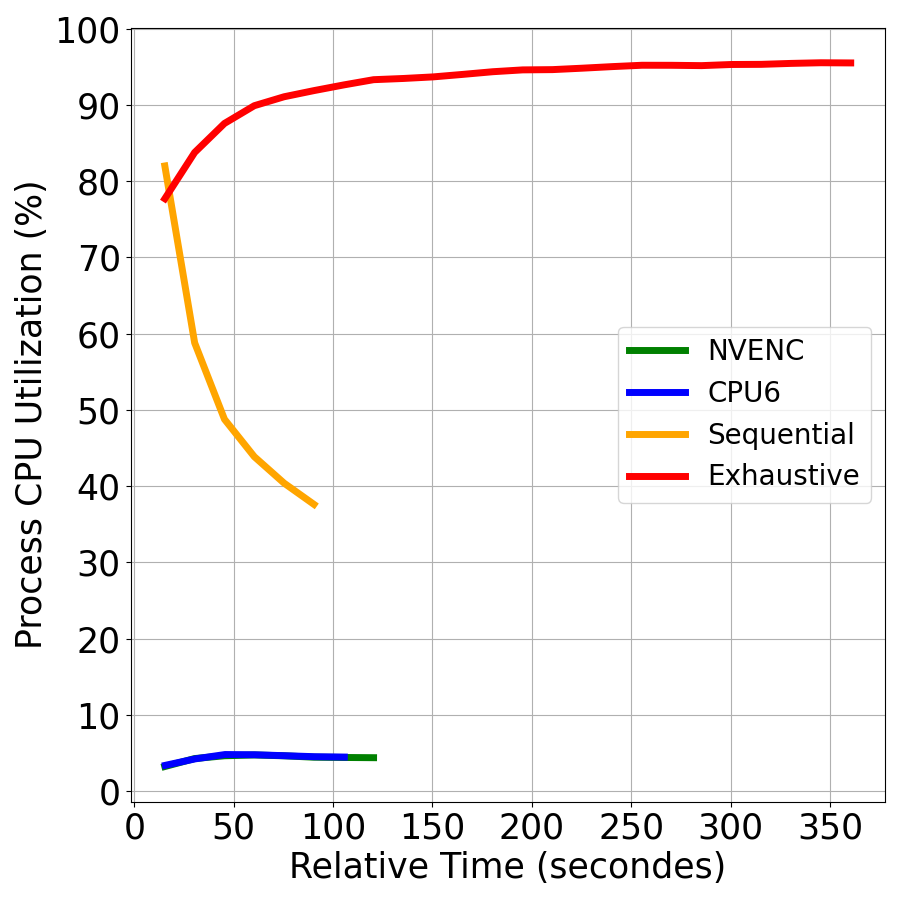}
    \caption{Paris-2: CPU}
    \label{fig:cpu-paris2}
\end{subfigure}\hfill
\begin{subfigure}[t]{0.22\textwidth}
    \centering
    \includegraphics[width=\linewidth]{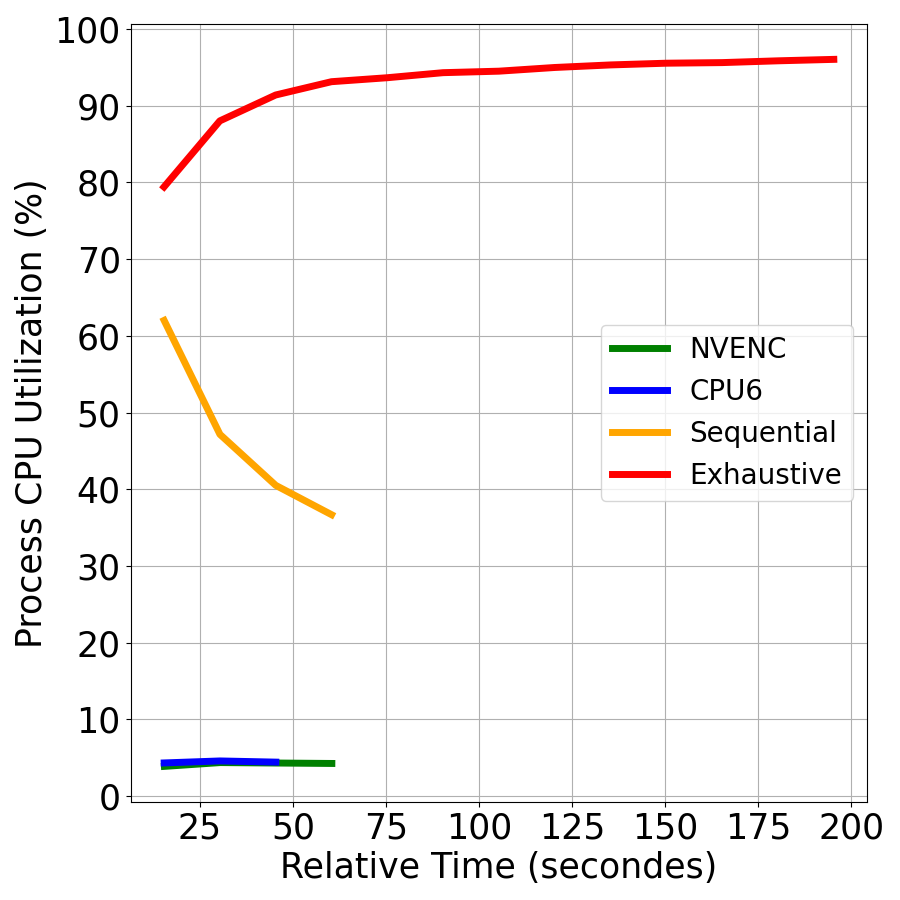}
    \caption{Gerrard Hall: CPU}
    \label{fig:cpu-gerrard}
\end{subfigure}\hfill
\begin{subfigure}[t]{0.22\textwidth}
    \centering
    \includegraphics[width=\linewidth]{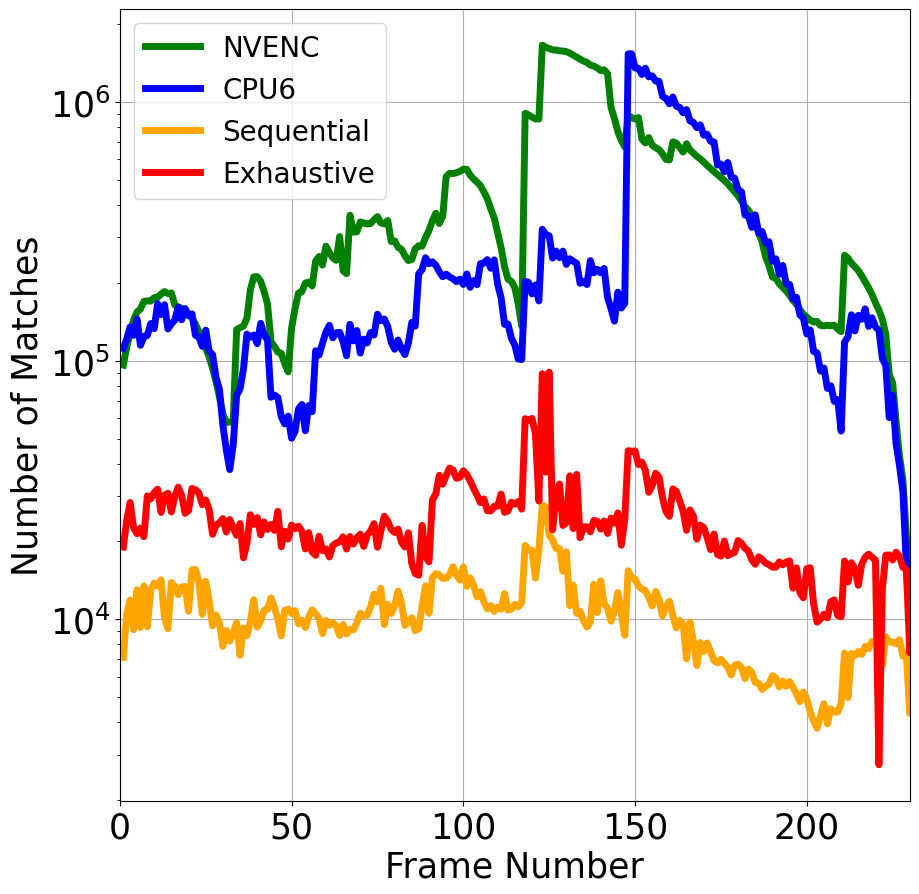}
    \caption{Per-image matches ($\times 10^5$)}
    \label{fig:number-matches}
\end{subfigure}
\caption{Process CPU Utilization (\%) for each method on (a) KITTI-00, (b) Paris-2, and (c) Gerrard Hall, and (d) the corresponding per-image match counts. Codec-based motion-vector pipelines (NVENC-AV1, AOM-AV1) produce higher match counts than classical feature pipelines under identical intrinsics and mapper settings.}
\label{fig:cpu-and-matches}
\end{figure*}

As shown in Table~\ref{results_table}, our \texttt{AOM AV1} method is over 3x faster than \texttt{SIFT Exhaustive} while using 95\% less CPU power. While \texttt{SIFT Sequential} has a comparable runtime, its CPU usage is an order of magnitude higher. This efficiency gap is visualized in Figure~\ref{fig:cpu-kitti}-\ref{fig:cpu-gerrard}.

In terms of geometric accuracy, our method consistently achieves the lowest median Sampson error (Table~\ref{sampson_distance}) and superior or competitive inlier ratios (Table~\ref{inlier_ratio}). Furthermore, our approach generates a denser set of matches than traditional methods (Figure~\ref{fig:number-matches}).

\begin{table}[t]
\caption{Efficiency of pre-stages only (Median over all sequences). Times in seconds, CPU is average during the pre-stages. The {\ul underlined value} is the best overall result.}
\centering
\resizebox{\columnwidth}{!}{%
\begin{tabular}{lccc}
\hline
\multicolumn{1}{c}{\textbf{Method}} &
  \multicolumn{1}{c}{\textbf{Pre-Processing (s)}} &
  \multicolumn{1}{c}{\textbf{Feature Matching (s)}} &
  \multicolumn{1}{c}{\textbf{CPU Usage (\%)}} \\ \hline
NVENC AV1 (our)  & 13.06 & 201.03 & {\ul 4.14}  \\
AOM AV1 (our) & {\ul 12.23} & 104.86  & 4.27  \\
SIFT Sequential   & 17.25 & {\ul 72.33} & 46.65 \\
SIFT Exhaustive   & 18.44 & 375.32 & 95.15 \\ \hline
\end{tabular}
}
\label{results_table}
\small
\end{table}

\begin{table}[t]
\caption{Median Sampson error per method per sequence. Lower is better. Our method using libaom av1 achieves the best overall.}
\centering
\resizebox{\columnwidth}{!}{%
\begin{tabular}{lcccc}
\hline
\multicolumn{1}{c}{\textbf{Seq.}} &
  \multicolumn{1}{c}{\textbf{NVENC AV1}} &
  \multicolumn{1}{c}{\textbf{AOM AV1}} &
  \multicolumn{1}{c}{\textbf{SIFT Sequential}} &
  \multicolumn{1}{c}{\textbf{SIFT Exhaustive}} \\ \hline
Dublin Seq 1 & {\ul 0.0003} & 0.0004 & 0.015 & 0.015  \\
Paris Seq 1 & 9.62E-05 & {\ul 5.82E-05} & 0.015 & N/A  \\
Paris Seq 2 & 1.44E-05 & {\ul 9.8E-06} & 0.083 & N/A \\
KITTI Seq 0 & {\ul 0.002} & {\ul 0.002} & 0.111 & N/A \\
Gerrard Hall & 0.004 & {\ul 0.003} & 0.005 & N/A \\
Person Hall & 0.003 & {\ul 0.001} & 0.022 & N/A \\ \hline
\end{tabular}
}
\label{sampson_distance}
\small
\end{table}

\begin{table}[t]
\caption{Median Inlier Ratio per method per sequence, higher is better. Our method using libaom on video sequences achieves  highly competitive inlier ratios compared to classical method.}
\centering
\resizebox{\columnwidth}{!}{%
\begin{tabular}{lcccc}
\multicolumn{1}{c}{\textbf{Seq.}} &
  \multicolumn{1}{c}{\textbf{NVENC AV1}} &
  \multicolumn{1}{c}{\textbf{AOM AV1}} &
  \multicolumn{1}{c}{\textbf{SIFT Sequential}} &
  \multicolumn{1}{c}{\textbf{SIFT Exhaustive}} \\ \hline
Dublin Seq 1 & {\ul 0.99} & 0.98& 0.96 & 0.51  \\
Paris Seq 1 & 0.96 & {\ul 0.99}  & 0.94 & 0  \\
Paris Seq 2 & 0.93 & {\ul 0.99} & 0.91 & 0 \\
KITTI Seq 0 & {\ul 0.98} & 0.95 & 0.96 & 0 \\
Gerrard Hall & 0.47 & {\ul 0.96} & 0.95 & 0 \\
Person Hall & 0.43 & 0.96 & {\ul 0.98} & 0 \\ \hline
\end{tabular}
}
\label{inlier_ratio}
\small
\end{table}

\section{Discussion and Limitations}

We observe that dense correspondence, while being beneficial for robustness and coverage, may significantly increase the computational cost of downstream bundle adjustment (BA) and reconstruction pipelines. This motivates future investigation into strategies to reduce the number of keypoints, therefore balancing geometric fidelity with efficiency.

The current implementation is also not optimized. Work will be done to reduce reading and writing to disk, reduce the number of loops, and parallelize the processing. In particular, the use of DAV1D decoder to extract metadata from AV1 bitstream might be explored.

A comprehensive Structure from Motion evaluation will be conducted. Reconstruction completeness (number of registered camera and triangulated points), Reprojection accuracy, 3D point cloud quality (using the Chamfer and Hausdorff distances to ground truth) and bundle adjustment scalability will be presented in a separate submission.

\section{Conclusion}

We have introduced a method leveraging AV1 motion vectors for dense, sub-pixel correspondence matching in Structure from Motion. Our approach achieves substantial speed and efficiency gains over SIFT-based baselines, while delivering competitive geometric accuracy and match density. The results highlight the potential of compressed-domain features for scalable, resource-efficient 3D vision. Future work will assess the impact on full SfM reconstruction and Bundle Adjustment performance.

\bibliographystyle{IEEEtran}
\bibliography{references}

\end{document}